\documentclass[a4paper,fleqn]{cas-dc}

\usepackage[numbers]{natbib}

\usepackage{amsfonts}
\usepackage{bbding}
\usepackage{latexsym}
\usepackage{graphicx}
\usepackage{fancyhdr}

\def\tsc#1{\csdef{#1}{\textsc{\lowercase{#1}}\xspace}}
\tsc{WGM}
\tsc{QE}
\tsc{EP}
\tsc{PMS}
\tsc{BEC}
\tsc{DE}

\begin{document}
\let\WriteBookmarks\relax
\def\floatpagepagefraction{1}
\def\textpagefraction{.001}

\title [mode = title]{Neural Radiance Fields with Torch Units}

\affiliation{organization={College of Computer Science and Technology},
            addressline={Zhejiang University},
            city={Hangzhou},
            state={Zhejiang},
            country={China}}

\author[]{Bingnan Ni}[style=chinese]

\author[]{Huanyu Wang}[style=chinese]

\author[]{Dongfeng Bai}[style=chinese]

\author[]{Minghe Weng}[style=chinese]

\author[]{Dexin Qi}[style=chinese]

\author[]{Weichao Qiu}[style=chinese]

\author[]{Bingbing Liu}[style=chinese]


\begin{abstract}
Neural Radiance Fields (NeRF) give rise to learning-based 3D reconstruction methods widely used in industrial applications. Although prevalent methods achieve considerable improvements in small-scale scenes, accomplishing reconstruction in complex and large-scale scenes is still challenging. First, the background in complex scenes shows a large variance among different views. Second, the current inference pattern, $i.e.$, a pixel only relies on an individual camera ray, fails to capture contextual information. To solve these problems, we propose to enlarge the ray perception field and build up the sample points interactions. In this paper, we design a novel inference pattern that encourages a single camera ray possessing more contextual information, and models the relationship among sample points on each camera ray. To hold contextual information,a camera ray in our proposed method can render a patch of pixels simultaneously. Moreover, we replace the MLP in neural radiance field models with distance-aware convolutions to enhance the feature propagation among sample points from the same camera ray. To summarize, as a torchlight, a ray in our proposed method achieves rendering a patch of image. Thus, we call the proposed method, Torch-NeRF. Extensive experiments on KITTI-360 and LLFF show that the Torch-NeRF exhibits excellent performance.
\end{abstract}


\maketitle
\thispagestyle{empty}
\section{Introduction}
Synthesizing novel photo-realistic views of a scene from several sparse input samples is widely used in industrial 3D reconstruction applications, $e.g.$, virtual reality, augmented reality, and autonomous driving simulation. Traditionally, computer graphics methods, such as meshes or voxel grids that store geometry and appearance information, are employed to solve these problems. Recently, learning-based methods gain signiﬁcant improvements thanks to the Neural Radiance Fields \cite{mildenhall2021nerf} (NeRF). A NeRF model is an MLP-based network that implicitly models the observed geometry and appearance by mapping a 5D input location (3D coordinates and 2D direction view) to volumes of opacity and color at that location. Subsequent works also extend the Neural Radiance Fields to various (indoor and outdoor) scenes \cite{martin2021nerf, reiser2021kilonerf, rematas2022urban, sitzmann2021light, yu2021pixelnerf, lu2023urban, cao2023scenerf} and try to speed up the inference phase \cite{deng2022depth, yan2023plenvdb, li2023steernerf, muller2022instant, neff2021donerf, rebain2021derf, chen2023mobilenerf} with explicit modelings by replacing the implicit mapping with sparse voxels.

\begin{figure*}[htbp]
	\includegraphics[scale=1]{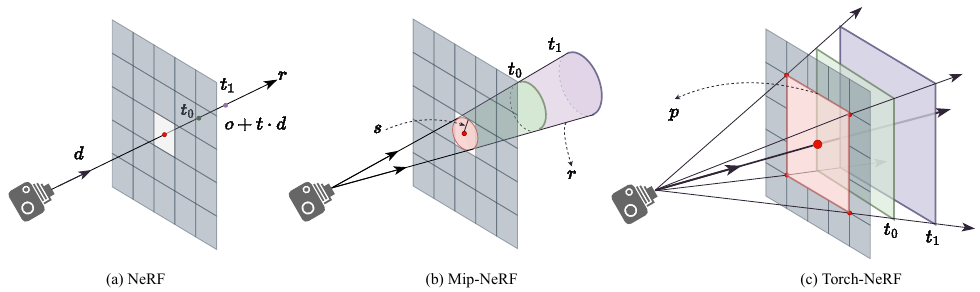}
	\caption{ Illustration of Ray Perception Field. (a) An individual camera ray in NeRF is only related to a single pixel. The camera ray $r$
goes through a pixel, thus, its perception field covers a single point. (b) Similarly, a camera ray in Mip-NeRF is also corresponding to a
single pixel. They utilize a conical ray to go through a circular area of a pixel. Thus, the perception field of such a ray is $\pi \cdot s^2$
, where s is a fixed value about half width of a pixel. (c) Different from the camera ray in prevalent methods which generates a single pixel, that in our method is able to render a patch of images of $p \times p$. Therefore, the ray perception field is $p \times p$ in our torch-NeRF.}
	\label{FIG:1}
\end{figure*}

Nevertheless, such approaches still cannot accomplish complex scenes well, especially in autonomous driving scenarios. First, different from synthesising novel views of a single object or scene, the background of the autonomous driving ones shows a huge variance between every two adjoint views. Second,the current inference pattern, $i.e.$, rendering a single pixel by an individual ray, make the synthesizing phase essentially degenerate into a mapping function. Since there is no interaction among samples, such an inference pattern is prone to noise values. Thus, ﬁtting isolated input locations without considering the relation among them is not appropriate in complex scenarios. To model an input camera ray with its neighborhood ones, a fundamental problem is how to enlarge the observation scope of rays. As shown in Fig. \ref{FIG:1}, a camera ray in the vanilla NeRF \cite{mildenhall2021nerf} only goes through a single point. Thus, the perception ﬁeld of rays in relevant neural radiance ﬁelds only covers a single point or a pixel. Similarly, the perception ﬁeld of rays in Mip-NeRF \cite{barron2021mip} is a circular area covering a single pixel. In this paper, we deﬁne such scope of observation of rays as Ray Perception Field and propose to enhance the relation- ship of a camera ray with its neighborhood ones.

Unlike the prevalent neural radiance ﬁeld frameworks, which project camera rays through the scene to generate a set of sample points and then process these sample points separately, we encourage a single camera ray possessing more contextual information and model the relationship among sample points from the same ray. To enlarge the ray perception ﬁeld, the models in our method synthesize a patch of pixels at the same time with 5D coordinates as input (location and viewing direction). First, we feed the input coordinates into a neural network to produce a patch of color and volume density. Then, we composite these values into a patch of pixels by using volume rendering techniques. As a result, the radiance ﬁeld model can learn and aggregate more contextual information for each camera ray.

Furthermore, the basic implementation of optimizing a radiance ﬁeld representation usually treats each sample point individually with MLP networks. However, such learning patterns fail to utilize the relationship among sample points from the same camera ray, resulting in noise space occupied value. We address this issue by introducing convolution along camera rays, which enables the interaction among sample points on a camera ray. Considering the distance between different sample points, we take it as weights and adaptively aggregate adjoint features.  In this way, the volume of these points would be smoothly distributed on each ray, thus effectively decreasing noise space occupancies and improving the quality of images.

As a whole, we call the proposed method torch-NeRF, $i.e.$, neural radiance ﬁelds with torch units, since the camera rays in our method are able to render a patch of image like a torch. The main contributions are summarized as follows:

\begin{itemize} 
\item We design a novel inference pattern for radiance field representation to enlarge the ray perception field so that a single camera ray could aggregate more contextual information during volume rendering. 
\item Considering sample points along the same camera ray, we build up their relationship with distance-aware convolutional components, making the volumes of these sample points smoothly distributed.
\item Extensive experiments show that the proposed method obtains significant improvement in large-scale scenarios. 
\end{itemize}

\section{Related Works}
In this section, we focus the literature review on radiance ﬁeld representation of 3D reconstruction and novel view synthesis tasks. The related works are concluded into three categories: techniques to improve the limitations of the neural radiance ﬁeld, methods to make the rendering of neural radiance ﬁelds customizable, and recent works to apply neural radiance ﬁelds to different applications.

\paragraph{Improved Technique on Radiance Field.} Aiming at improving the neural radiance ﬁeld, previous works mainly focus on three limitations of radiance ﬁeld representation.

First, considering the computational cost of NeRF network, methods \cite{deng2022depth, garbin2021fastnerf, li2023steernerf, muller2022instant, rebain2021derf, reiser2021kilonerf, sitzmann2021light, chen2023mobilenerf, yan2023plenvdb} are proposed to accelerate the inference phase. Speciﬁcally, PlenVDB \cite{yan2023plenvdb} directly learns scenario information on VDB data structures, and Mobile-NeRF \cite{chen2023mobilenerf} enables real-time neural rendering on mobile devices. Besides, DS-NeRF \cite{deng2022depth} attempts to utilize the sparse output from Structure from Motion to regularize the NeRF to speed up the optimization. AutoInt \cite{lindell2021autoint} assemble the integral phase in the network.  Recently, NVIDIA proposes the Instance-NGP \cite{muller2022instant} and achieves signiﬁcant improvement in inference speed. Second, several works \cite{trevithick2020grf, wang2021ibrnet, yu2021pixelnerf, zhang2020nerf++, chen2023gm} try to explore generalization and transferability. They propose to combine image features with NeRF to transfer geometry prior to the target scene. For instances, GRF \cite{trevithick2020grf} learns and projects local features of each pixel to the radiance ﬁeld, which achieves better-generalized representation ability. PixelNeRF \cite{yu2021pixelnerf} is another representative method that proposes a CNN-based encoder to extract the image features under few-shot scenarios. Besides, chen $et al.$ \cite{chen2023gm} propose a geometry-guided attention mechanism to register the appearance code from multi-view 2D images to a geometry proxy which can alleviate the misalignment between inaccurate geometry prior and pixel space. Third, to extend the neural radiance ﬁeld into real-world scenarios, researchers propose works \cite{kundu2022panoptic, martin2021nerf, rematas2022urban, tancik2022block, xian2021space} to learn and reconstruct complex outdoor scenes, especially in autonomous driving and street scenes.  Among these works, Block-NeRF \cite{tancik2022block} decouples and learns different streets with a separated model and aligns the borders of different NeRF models. PNF \cite{kundu2022panoptic} introduces the deﬁnition of panoptic segmentation and trains an individual NeRF for an object.

\textit{Key Difference.} Different from the aforementioned methods who propose to improve the limitations of NeRFs, we design a novel inference pattern to improve the learning of contextual information and adapt to complex scenes.

\paragraph{Conditional Neural Radiance Field.} Another line of work is to make the generated images customizable. Although the neural radiance ﬁeld achieves to reconstruct the geometry and colors of scenes, it does not allow us to customize or control the shapes or colors of the scenes.

Similar to Condition-GAN \cite{mirza2014conditional} and Style-GAN  \cite{karras2020analyzing}, this line of works \cite{chan2021pi, liu2021editing, muller2022autorf, niemeyer2021giraffe, pumarola2021d, rematas2021sharf, schwarz2020graf, yang2021learning} usually using an extra embedding to model the appearance. As a seed work, EditNeRF \cite{liu2021editing} proposes to propagate coarse 2D scribbles from users to the 3D space and modify the color or shape of the local region.  Subsequent works further borrow ideas from GANs and achieve considerable improvement. As a representative, GRAF \cite{schwarz2020graf} composes a generator with a discriminator and then proposes a voxel representation to explicitly generate 3D features.  Following, GIRAFFE \cite{niemeyer2021giraffe} attempts to separate the objects from the background and models the shape and appearance embeddings for individual object with a single neural feature ﬁeld. Besides, neural radiance ﬁelds are also applied to digital human body \cite{corona2022lisa, gafni2021dynamic, liu2021neural, peng2021animatable, weng2022humannerf, yen2021inerf}. They attempt to reconstruct the human face and body by integrating radiance ﬁeld models with traditional methods. For example, 4D Facial Avatar \cite{gafni2021dynamic} proposes to combine 3DMM and NeRF models together to achieve dynamic object modeling in a neural radiation ﬁeld. With an input video, it can realize editing the positions and expressions of faces.  Besides, Animatable \cite{peng2021animatable} introduces neural radiance ﬁelds to generate a deformation ﬁeld and enables human body modeling. Given a gauge space or a template, it estimates the gauge space from different observation spaces.

\paragraph{Neural Radiance Fields in Different Applications.} In addition  to the aforementioned works, attaching NeRFs with text, video, and medical ﬁelds also achieve excellent performance in different applications.

As an implicit representation, the neural radiance ﬁeld provides novel means for traditional image processing methods \cite{chen2021learning, czerkawski2021neural, liu2022nerf, wang2022nerf}. That is, reconstruct the 3D geometry priors with the radiance ﬁeld and then combine the prior model with prelevant compression, denoising, super-resolution,and in-painting methods. For an instance, Knitworks \cite{czerkawski2021neural} proposes a framework for image representation, which implements image synthesis by optimizing the distribution of image patches in an adversarial manner. Moreover, exploring NeRFs with other modes of information, such as text and audio, is another popular research area. For example, CLIP-NeRF \cite{baatz2022nerf, wang2022clip} combines a CLIP-Driven manipulation module with a disentangled conditional NeRF. Therefore, the generation of NeRF model is controlled by the input text. Besides, applying NeRFs \cite{kasten2021layered, li2021neural, peng2021neural, xian2021space} on video compressing, editing, and generation are also ongoing.  In which, Layered Neural Atlases \cite{kasten2021layered} is a representative work that unfolds input videos into several slices and stores the slices with a group of parameters. Moreover, the neural radiance ﬁelds are also applied to robotics \cite{li20223d, adamkiewicz2022vision} and medical image processing \cite{shen2022nerp} applications.

\begin{figure*}[htbp]
	\includegraphics[scale=.95]{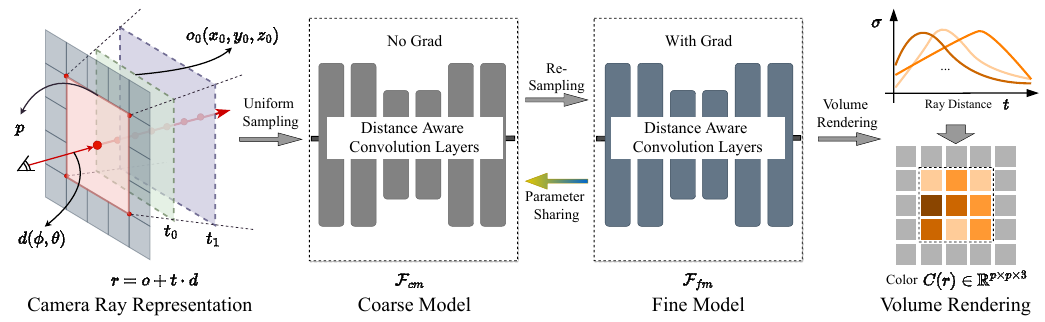}
	\caption{The whole framework of the proposed method. Similar to NeRF, the input coordinates compose of a 3D position location and a 2D view direction. First, we input uniform sampled points along each ray to a coarse model. The coarse model only outputs the density of each sample point. Second, we re-sample points according to the output density of the coarse model. Next, the re-sampled points are input into a fine model. Finally, we render a patch of pixels $p \times p$ with the input ray. It is worth noting that the coarse model is training free.}
	\label{FIG:2}
\end{figure*}

\begin{figure}[htbp]
	\includegraphics[scale=.35]{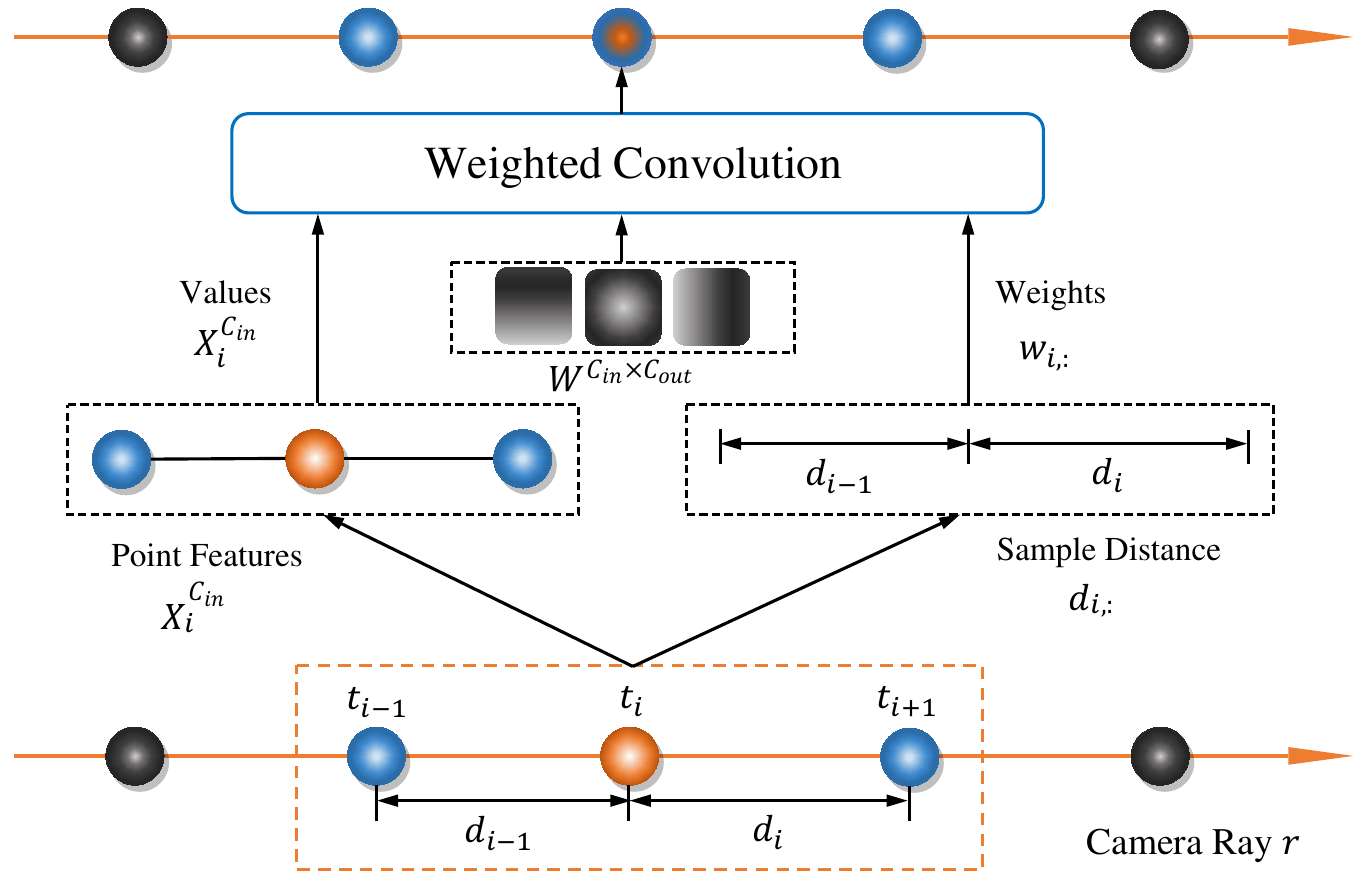}
	\caption{ Illustration of Distance aware Convolution Operation along Rays. We conduct a weighted 1D convolution operation along ray with a sliding window over several sample points.}
	\label{FIG:3}
\end{figure}

\section{Methodology}
\begin{figure*}[htbp]
	\includegraphics[scale=.7]{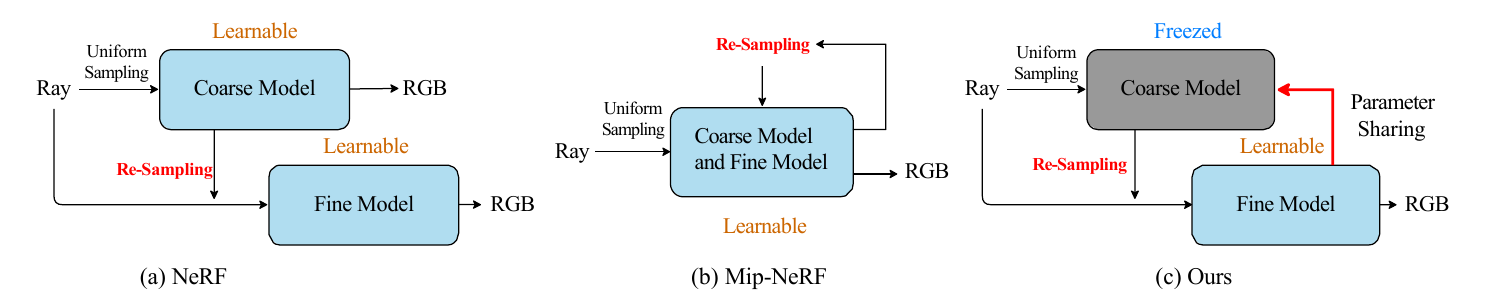}
	\caption{ Illustration of the proposed network structure. (a) The vanilla Neural Radiance Field method\cite{mildenhall2021nerf} trains a coarse model and a fine model independently. (b) The Mip-NeRF\cite{barron2021mip} only uses a single model and backward it twice. (c) In our method, we only train the fine model and update the coarse model with the updating of the fine model.}
	\label{FIG:4}
\end{figure*}

In this section, we illustrate the proposed method, Neural Radiance Field with Torch Units in detail. First, we demonstrate the whole framework and details of patch-wise volume rendering pattern as shown in Fig. \ref{FIG:2}. Second, we introduce the proposed distance aware convolution along rays as shown in Fig. \ref{FIG:3}. Next, we present the optimization strategy and objective functions of the whole network in Fig. \ref{FIG:4}.

\subsection{Neural Radiance Field With Torch Units.}
\paragraph{Neural Radiance Field.} Our proposed method is based on neural radiance field \cite{mildenhall2021nerf}, which is to learn a mapping function to project a 5D coordinate of an input point to a RGB color and corresponding space occupancy. Formally, considering a camera ray r, we sample N sample points along the camera ray by $r=o+t \cdot d$, where $t \in [1, N]$ is the location of a sample point on the camera ray. Denote the position and direction of the $n$-th input sample point on the camera ray, as $o_n = (x_n, y_n, z_n)$ and $d = (\phi, \theta)$ in the world coordinate. With an MLP network F, the location and direction, $i.e.$, 5D coordinate, is projected into a RGB color $c_n$ and space occupancy $\sigma_n$ as $(c_n, \sigma_n) = F(x_n, y_n, z_n, \phi, \theta)$. Through collecting the output RGB colors $(c_1,... , c_N )$ and occupancies $(\sigma_1, ... , \sigma_N )$ of each sample points on the camera ray, the expected color $C(r)$ of the camera ray is rendered by
\begin{equation} \label{Eq1}
\begin{aligned}
    C(r)=\sum_{n=1}^N{(1-exp(-\sigma_n(t_{n+1}-t_n)))}c_n,\\
    where\ T_n=exp(-\sum_{m=1}^{n-1}\sigma_m(t_{m+1}-t_m))
\end{aligned}
\end{equation}
As a result, the color of the pixel, which is gone through by the camera ray $r$, is also represented by $C(r)$.

\paragraph{Enlarge the Ray Perception Field.} In the above method, the perception field of each input camera ray only covers a single pixel, since a individual ray is regularized by a single pixel. In comparison, we propose to enlarge the ray perception field by rendering a patch of pixels $p\times p$ with a single input camera ray $r$. To render a patch of pixels at the same time, sample points on a camera ray are also able to model a patch of pixels. Therefore, the inference of the $n$-th sample point is
\begin{equation}
\begin{aligned}
    (c_n, \sigma_n)=F(x_n, y_n, z_n, \phi, \theta),
\end{aligned}
\end{equation}
where $c_n \in \mathbb{R}^{p \times p \times 3}$, $\sigma_n \in \mathbb{R}^{p \times p \times 1}$ is the output color and space occupancy of the $n$-th sample point on the camera ray $r$. Then, with the respective color and space occupancy of sample points along the ray, a patch image would be rendered following Eq. (\ref{Eq1}).

\subsection{Distance Aware Convolution Along Rays.}
In this part, we illustrate the proposed distance aware convolution along rays. Prevalent neural radiance field methods usually process individual sample points separately, which would lead to noisy values along the same ray, especially on space occupancy. For this, Mip-NeRF360 \cite{barron2022mip} attempts to regularize the space occupancies $\sigma$ directly with a distillation loss. However, simply regularizing the volumes still ignore the relationship modeling among samples points along rays. In comparison, we explicitly model the relationship among sample points by utilizing distance aware convolutions to enhance the feature interaction and decrease the noise occupancy values.

\paragraph{Convolution Along Rays.} We first briefly formulate the standard 1D convolution and basic operation. Then transfer it to the distance aware convolution. For convenience, we only focus on the anchor sample point in the convolution, though there are N sample points $(o_1, ... , o_N )$ along the camera ray $r$. Initially, the input feature of the anchor sample point is denoted as $X \in \mathbb{R}^{C_{in}}$, where $C_{in}$ is the channel number. Let $Y \in \mathbb{R}_K^{C_{in} \times C_{out}}$ be the output feature and $W \in \mathbb{R}_K^{C_{in} \times C_{out}}$ be the convolutional filters. With a standard 1D-convolution operation, the corresponding calculation of the anchor sample point is
\begin{equation}
\begin{aligned}
    Y^{C_{out}}=\sum_{k=1}^K{X^{C_{in}} \odot W_k^{C_{in} \times C_{out}}},
\end{aligned}
\end{equation}
where $K$ is the size of 1D-convolution kernel and $\odot$ is the convolution operation.

\paragraph{Weighted Convolution.} Following the above formulation, we model the relationship among these sample points. Owing to the sampling strategy in radiance field, the sample points are not always uniformly distributed along rays. Therefore, we take the effects of the distance between each other into consideration as shown in Fig. \ref{FIG:3}.

When the network processes the anchor sample point, $K$ sample points, $i.e.$, $(t_1, ... , t_K)$, are included in the convolutional operation. In principle, we expect that the closer the distance between a sample point to the anchor one, the stronger the connection should be. Therefore, we calculate the weight of each sample point to the anchor one based on the distance between samples. First, the distance is calculated according to the input location coordinates as

\begin{equation}
\begin{aligned}
    d_k=\|t_{anchor}-t_k\|,\ where\ k \in [1,K]
\end{aligned}
\end{equation}
In which, the $\|.\|$ is the $L^1$ norm and $t_anchor$ is the location of the anchor point. Moreover, we normalize the distance, $i.e.$, $d$, to a range from 0 to $\pi$/2, and then apply cosine function as an activation function over the normalized distance value. The weight from the $k$-th item in the convolution kernel to the anchor sample point is defined as
\begin{equation}
\begin{aligned}
    w_k=cos(norm(d_k)),\ where\ k \in [1,K]
\end{aligned}
\end{equation}
In this way, the whole formulation of distance aware convolution operation is
\begin{equation}
\begin{aligned}
    Y^{C_{out}}=\sum_{k=1}^K{(X^{C_{in}} \odot W_k^{C_{in}\times C_{out}}) \times w_k},
\end{aligned}
\end{equation}
where $k$ is the sample index in the convolution kernel $W_K$.

\subsection{Network and Optimization.}
In this part, we illustrate the network design and optimization of the proposed method. First, we introduce the design and updating strategy of the modules. Next, we show the objective functions to optimize the network parameters.

\paragraph{Network Structure}. In terms of the network structure, we propose a simple yet effective strategy to reduce the training overhead. Prevalent neural radiance field methods mainly contains two sampling phases. They input uniformly sampled points to a coarse model and estimate the weights of them. According to the weights, they re-sample a series of points and inputs to a fine model. Essentially, the coarse and fine models are optimized separately without gradient-flow.

As shown in Fig. 4a, there is no gradient passing between the coarse and fine model in the vanilla NeRF. In other words, a ray should be processed twice through the coarse and fine models, respectively. Moreover, Mip-NeRF \cite{barron2021mip} takes a different updating strategy, which uses a shared model to process sampled points during uniform sampling and re-sampling as shown in Fig. 4b. However, although a shared model effectively reduces the storage overhead, sample points from uniform sampling and re-sampling are still required to be optimized altogether in a training iteration. In comparison, our proposed framework contains two modules, $i.e.$, a coarse model, $F_{cm}$, and a fine model, $F_{fm}$. The coarse model is training free and only the fine model is required to be trained. The parameters in coarse model are accordingly adjusted with the updating of the fine model every $M$ iteration as shown in Fig. 4c.

\paragraph{Objective Function.} In this way, we optimize the learnable parameters, i.e., $F_{fm}$, with the loss function as
\begin{equation}
\begin{aligned}
    \mathcal{L}_{total}=\mathcal{L}_{mse} + \mathcal{L}_{ssim}
\end{aligned}
\end{equation}
where $\mathcal{L}_{ssim}$ is the structure similarity loss, and $\mathcal{L}_{mse}$ is the MSE loss function. It is worth noting that the above loss functions are calculated in a patch of image $p \times p$, which is rendered by a single camera ray. Unlike traditional MSE loss, we set different weights for each pixel, so that the network can focus on the center point and learn the correlation between different pixels. So we define the $\mathcal{L}_{mse}$ as
\begin{equation}
\begin{split}
    \mathcal{L}_{mse}&=\frac{1}{p \times p}  \sum\nolimits_{i}^{p} \sum\nolimits_{j}^{p}{w_{i,j} \cdot ||\hat{C}_f(i,j)-C(i,j)||_2^2} \\
    w_{i,j}&=exp(-||(i,j)-(i_c,j_c)||_2)
\end{split}
\end{equation}
where $C(i,j)$ and $\hat{C}_f(i,j)$ are the ground truth and fine volume predicted RGB colors for ray $r$ in a patch respectively. $(i,j)$ and $(i_c,j_c)$ are the coordinate of a pixel and the coordinate of the center pixel in the patch respectively.

\section{Experiments}
In this section, we conduct experiments to show the effectiveness of the proposed method. First, we introduce the experimental settings and implementation details. Second, we show the quantitative comparisons with other related works. Next, we analysis the pros and cons of our proposed method through qualitative comparisons. Finally, we present ablation studies on our proposed methods.

\subsection{Experimental Settings.}
\paragraph{Datasets.} We evaluate the proposed torch-NeRF on two widely used datasets with complex backgrounds, $i.e.$, KITTI-360 \cite{barron2022mip} and LLFF \cite{mildenhall2019local}. The KITTI-360 dataset involves images taken by forward-facing cameras in complex outdoor dynamic scenes. It is the first benchmark that evaluates the synthesizing of color and appearance images from novel views. KITTI-360 contains 5 individual sequences of images with a resolution of 1408$\times$376, and each sequence corresponds to a continuous driving trajectory. Besides, LLFF is a series of scenes of a single object or area with complex background. It includes eight indoor scenes or outdoor scenes. Different from images in KITTI-360, which are collected by the forward-facing camera, those in the LLFF dataset are collected by the multi-views camera.

\paragraph{Evaluation.} Following the evaluation metrics in \cite{barron2021mip, barron2022mip, kundu2022panoptic, mildenhall2021nerf, sun2022direct}, we present quantitative and qualitative results by showing the generated images of novel views and reporting the mean PSNR, SSIM \cite{wang2004image}, and LPIPS \cite{zhang2018unreasonable} across the test images from different scenes in respective datasets. Specifically, The SSIM stands for the structural similarity between the generated image and the ground truth image. As the range of SSIM is from -1 to 1, where 1 indicates the best performance and -1 indicates the worst. The LPIPS stands for the perceptual image patch similarity, which is calculated to judge the perceptual similarity between features.

\paragraph{Implementation.} In our method, the whole network contains a coarse module and a fine module. The coarse and fine modules are of the same network structures, which respectively include eight distance aware convolution layers followed by ReLU activations and batch-wise normalizations. The channels of the convolution layers are set to 128, 256 and 512 in ascending order. To achieve rendering a patch of pixels at once, the final output channel of the network is $p \times p \times 3$. For the best performance, we set $p$ to 5 in experiments of KITTI-360 and LLFF. The size of 1D-convolutional kernels K in distance aware convolution is set to 3. At training time, we set the number of sample points N to 512 on KITTI-360 and LLFF. In terms of the optimization strategy, the fine model is updated by backpropagation and the coarse model is updated every 200 iterations by passing parameters from the fine model.In the inference phase, we only use the center point of a patch to predict the RGB color of the pixel.

\begin{table*}
\caption{ Performance comparisons with other methods on KITTI-360 \cite{liao2022kitti}. KITTI-360 contains 5 individual sequences. The results are evaluated on average of PSNR, SSIM, and LPIPS among all sequences. NeRF-Based refers to neural radiance based method. Require Priors refers to require semantic and instance labels. Multi-Scale refers to require a multi-scale modeling.}\label{tbl1}
\rmfamily
\resizebox{\textwidth}{!}{
\begin{tabular}{l|ccc|ccc}
\toprule
Methods & NeRF-Based & Require Priors & Multi-Scale &PSNR $\uparrow$ &SSIM $\uparrow$ &LPIPS $\downarrow$ \\
\midrule
FVS \cite{riegler2020free} + PSPNet \cite{zhao2017pyramid} & \XSolidBrush & \XSolidBrush & \XSolidBrush & 20.00 & 0.790 & 0.193 \\
PBNR \cite{kopanas2021point} + PSPNet \cite{zhao2017pyramid}& \XSolidBrush & \XSolidBrush & \XSolidBrush & 19.91 & 0.811 & \textbf{0.191} \\
PCL \cite{lindell2021autoint} + PSPNet \cite{zhao2017pyramid} & \XSolidBrush & \XSolidBrush & \XSolidBrush & 12.81 & 0.576 & 0.549\\
\hline
NeRF \cite{mildenhall2021nerf} & \checkmark & \XSolidBrush & \XSolidBrush & 21.18 & 0.779 & 0.343 \\
Mip-NeRF \cite{barron2021mip} & \checkmark & \XSolidBrush & \checkmark & 21.54 & 0.778 & 0.365 \\
PNF \cite{kundu2022panoptic} & \checkmark & \checkmark & \XSolidBrush & 22.07 & 0.820 & 0.221\\
\hline
Ours & \checkmark & \XSolidBrush & \XSolidBrush & \textbf{22.25} & \textbf{0.827} & 0.248\\
\bottomrule
\end{tabular}
}
\end{table*}

\begin{table}
\caption{Performance comparisons with other methods
on LLFF \cite{mildenhall2019local}. LLFF contains 8 different scenes. The results are
evaluated on average of PSNR, SSIM, and LPIPS on all scenes.}\label{tbl2}
\rmfamily
\resizebox{\tblwidth}{!}{
\begin{tabular}{l|ccc}
\toprule
Methods &PSNR $\uparrow$ &SSIM $\uparrow$ &LPIPS $\downarrow$ \\
\midrule
NeRF \cite{mildenhall2021nerf} & 26.85 & 0.826 & 0.226 \\
Plenoxels \cite{fridovich2022plenoxels}& 26.41 & 0.831 & 0.218 \\
DVGO \cite{sun2022direct} & 26.60 & 0.837 & 0.205 \\
Mip-NeRF \cite{barron2021mip}& 26.80 & 0.823 & 0.230 \\
Mip-NeRF360 \cite{barron2022mip} & 26.63 & 0.835 & \textbf{0.196} \\ 
\hline
Ours & \textbf{27.01} & \textbf{0.839}& 0.215\\
\bottomrule
\end{tabular}
}
\end{table}

\subsection{Quantitative Comparison.}
We illustrate the advantages of neural radiance field with
torch units by comparing it with other widely used methods
on different datasets. First, we compare torch-NeRF with
others on KITTI-360. Then, we show the comparison of
torch-NeRF with related methods on LLFF.

\paragraph{Performance on KITTI-360 Dataset.} We conduct comparisons of our work with other methods. Following the home page of KITTI-360 \cite{liao2022kitti} benchmark, we compare with the following methods, including FVS \cite{riegler2020free}, PBNR \cite{kopanas2021point}, PCL \cite{liao2022kitti}, NeRF \cite{mildenhall2021nerf}, Mip-NeRF \cite{barron2021mip}, and PNF \cite{kundu2022panoptic}. Among these methods, FVS, PBNR, and PCL are implemented based on PSPNet \cite{zhao2017pyramid}. As shown in Tab. \ref{tbl1}, the PBNR achieves the best perceptual performance of 0.191 LPIPS and a considerable PSNR result of 19.91. Meanwhile, FVS achieves 20.00 PSNR with 0.790 structure similarity.

Moreover, NeRF \cite{mildenhall2021nerf}, Mip-NeRF \cite{barron2021mip}, and PNF \cite{kundu2022panoptic} are neural radiance field-based methods, which perform better in terms of PSNR and SSIM. NeRF is the baseline of this line of methods, which improves the performance to 21.18 PSNR, 0.779 SSIM. Based on NeRF, Mip-NeRF uses a 3D conical frustum-defined camera ray and encodes the camera with an integrated positional encoding. As a result, it achieves 21.54 PSNR with 0.778 SSIM. The PNF proposes a different way that applying the semantic and instance priors. Besides, it also introduces the thinking from panoptic segmentation to reconstruction by modeling an individual NeRF for each object and a shared one to the stuff. In this way, it achieves 22.07 PSNR with 0.820 SSIM. Different from PNF, which requires extra semantic and instance labels to improve the performance, our method achieves results of 22.25 PSNR with 0.827 SSIM without semantic or instance prior information.

\paragraph{Performance on LLFF Dataset.} Next, we compare the torch-NeRF with NeRF \cite{mildenhall2021nerf}, Plenoxels \cite{fridovich2022plenoxels}, DVGO \cite{sun2022direct}, Mip-NeRF \cite{barron2021mip}, and Mip-NeRF360 \cite{barron2022mip} on the LLFF \cite{mildenhall2019local} dataset as shown in Tab. \ref{tbl2}. Among these methods, Plenoxels exploits the spherical harmonics coefficient to represent the sparse grid voxel representation and achieves 26.41 PSNR with 0.831 SSIM. The Mip-NeRF360 extends the Mip-NeRF to unbounded scenes to deal with the real-world arbitrary camera pose orientation and efficiently generates volumetric density to model the unbounded 360 scenes. It obtains a performance of 26.63 PSNR with 0.835 SSIM. Moreover, our method surpasses others by a large margin in terms of PSNR and SSIM. We achieve the performance of 27.01 PSNR and 0.839 SSIM. In comparison, our method wins about 0.16 PSNR and 0.013 SSIM than the vanilla NeRF, which is the baseline of our method.

\begin{figure*}[htbp]
	\includegraphics[scale=1]{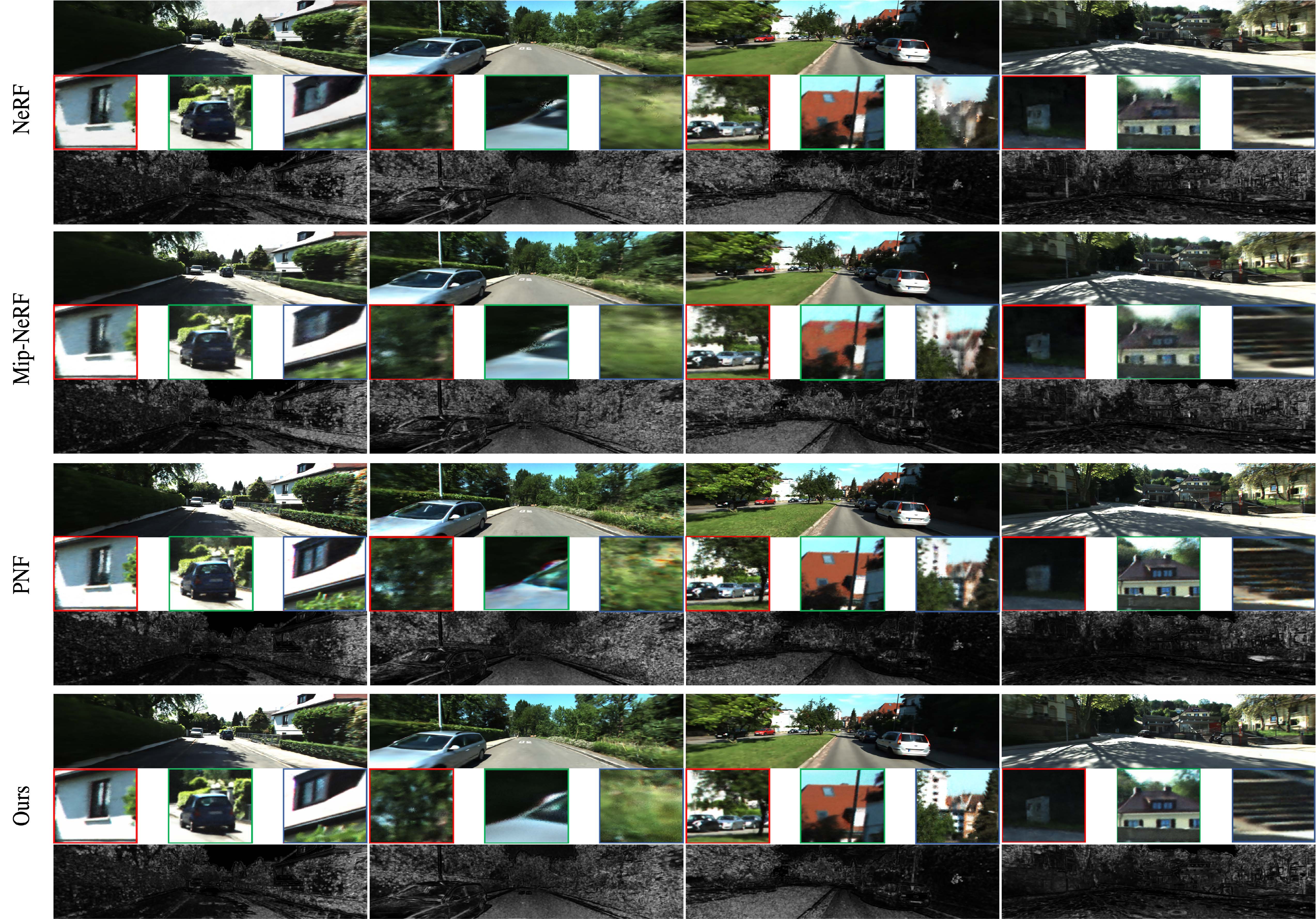}
	\caption{Visualization of synthesized images and error images. We compare with the other three neural radiance-based methods by referring to the project page of KITTI-360. The error images are generated according to (0.5 - SSIM/2). Thus, the bright region means a large error and the dark one means the low error. To make the contrast more obvious, we zoom-in on three patches of images from each synthesized one.}
	\label{FIG:5}
\end{figure*}

\subsection{Qualitative Analysis.}
We show the qualitative visualization of the proposed method compared with related works. We present synthesized samples and error images of NeRF \cite{mildenhall2021nerf}, Mip-NeRF, and PNF in Fig. \ref{FIG:5}. First of all, we should clarify that NeRF \cite{mildenhall2021nerf}, Mip-NeRF, and our proposed torch-NeRF are trained without any priors. In contrast, the PNF requires semantic and instance masks during optimization. Nonetheless, the performance of our proposed method still exceeds the PNF in the metrics. In the following part, we highlight the advantages qualitatively of our proposed method, which models the neural radiance field with torch units.

Considering the advantages, first, the torch-NeRF performs better on the overall structure of images. As presented in the last image, our method can learn contextual information with a large ray perception field. Next, our method also performs better on noisy colors under sunlight. As shown in the second column of the images, PNF shows distinct noise faculae. In comparison, ours can solve the reflection effect. Finally, our proposed method could handle the edge of the scene better. As illustrated in Fig. \ref{FIG:5}, objects on the edge are blurred and fuzzy. However, images in ours can preserve the shape of those objects.

\begin{table}[htbp]
\caption{Ablation study on different modules in our proposed
method on LLFF. The \textit{Dis.} stands for using \textit{Distance Aware Convolution} instead of \textit{MLP.} The \textit{En.} stands for \textit{Enlarge Ray Perception Field.} The $\mathcal{L}_{ssim}$ stands for the structure similarity loss.}\label{tbl3}
\rmfamily
\resizebox{\tblwidth}{!}{
\begin{tabular}{l|ccc|ccc}
\toprule
Methods & \textit{Dis.}&\textit{En.}&$\mathcal{L}_{ssim}$&PSNR $\uparrow$ &SSIM $\uparrow$ &LPIPS $\downarrow$ \\
\midrule
EXP-1 & - & - & - & 26.85 & 0.826 & 0.226 \\
EXP-2& \checkmark & - & - & 26.91 & 0.829 & 0.232 \\
EXP-3 & \checkmark & \checkmark & - & 26.99 & 0.818 & 0.227 \\
\hline
Ours & \checkmark & \checkmark & \checkmark & \textbf{27.01} & \textbf{0.839} & \textbf{0.215}\\
\bottomrule
\end{tabular}
}
\end{table}

\begin{table}[htbp]
\caption{Ablation study on the size \textit{p} of the Ray Perception Field. The experiments are conducted based on LLFF.}\label{tbl4}
\rmfamily
\resizebox{\tblwidth}{!}{
\begin{tabular}{l|c|ccc}
\toprule
Methods & \textit{Per. Fi.}&PSNR $\uparrow$ &SSIM $\uparrow$ &LPIPS $\downarrow$ \\
\midrule
EXP-1 & $1 \times 1$ & 26.91 & 0.829 & 0.232 \\
EXP-2& $3 \times 3$ & 26.95 & 0.827 & 0.221 \\
EXP-3 & $5 \times 5$ & \textbf{27.01} & 0.839 & \textbf{0.215} \\
EXP-4 & $7 \times 7$ & 26.49 & 0.841 & 0.228 \\
EXP-5 & $9 \times 9$ & 26.44 & \textbf{0.845} & 0.246 \\
\bottomrule
\end{tabular}
}
\end{table}

\begin{table}[htbp]
\caption{Ablation study on different updating strategies on LLFF. The baseline is without our modules. Others adopt a $5 \times 5$ ray perception field and distance aware convolution. \textit{Sep.} refers to optimize coarse and fine models separately. \textit{Sha.} refers to share the two models. And ours refers to the proposed optimizing strategy.}\label{tbl5}
\rmfamily
\resizebox{\tblwidth}{!}{
\begin{tabular}{l|c|ccc}
\toprule
Methods & Strategy &PSNR $\uparrow$ &SSIM $\uparrow$ &LPIPS $\downarrow$ \\
\midrule
Baseline & \textit{Sep.} & 26.85 & 0.826 & 0.226 \\
EXP-1& \textit{Sep.} & 26.89 & 0.834 & \textbf{0.213} \\
EXP-2 & \textit{Sha.} & 26.61 & 0.816 & 0.231 \\
\hline
EXP-3 & Ours & \textbf{27.01} & \textbf{0.839} & 0.215 \\
\bottomrule
\end{tabular}
}
\end{table}

\subsection{Ablation Studies.}
In this part, we first evaluate the effectiveness of each module, $i.e.$, inference pattern to enlarge the ray perception field, distance aware convolution along rays, and the structural similarity loss. Second, we analysis on the size of the ray perception field. Next, we study the different optimization strategies for updating the coarse and fine models.\\
\paragraph{Effectiveness of the Proposals.} We conduct ablation studies on different modules in the proposed method. It is worth noting that the network of baseline neural radiance field is composed of MLP. As shown in Tab. \ref{tbl3}, with distance aware convolution, using distance aware convolution based on the baseline surpasses the baseline method that uses the MLP layers, about 0.06 and 0.003 on PSNR and SSIM, respectively. Then, the PSNR again improve by about 0.08, when we enlarge the perception field of the camera ray. When using SSIM loss, PSNR only improved slightly, but SSIM improved by 0.021, indicating that adding SSIM loss to total loss is effective.

\paragraph{Ray Perception Field.} Moreover, we analyze the size of the ray perception field by enlarging the size from $3 \times 3$ to $9 \times 9$. As shown in Tab. \ref{tbl4}, with the expansion of the size of the ray perception field, the structure similarity shows an improving tendency accordingly. However, the performance of PSNR is not distributed consistently with the improvement of SSIM. Especially, the PSNR performance is increased from 26.91 to 27.01, when $p$ expands from 1 to 5. Then, it decreases to 26.44 PSNR, when $p$ is set to 9.This shows that the bigger the receptive field is, not the better. When $p=5$, a relatively good comprehensive result is produced.

\paragraph{Optimization Strategy.} Next, we study the different updating strategies to optimize the coarse and fine models. We try three different strategies: using separated coarse and fine models, sharing the same network between coarse and fine models, and updating the coarse model by passing parameters from the fine model. As shown in Tab. \ref{tbl5}, the baseline is conducted with MLP-based network and the origin rays. Other experiments are based on $5 \times 5$ ray perception field and distance aware convolution. The EXP-1 achieves 26.89 PSNR with 0.834 SSIM. Similar to the strategy in Mip-NeRF, the EXP-2 achieves 26.61 PSNR with 0.816 SSIM. Although we utilize a separate coarse model and fine model in our method, the parameters in the coarse one is updated with the optimization of the fine model.

\subsection{Limitations.}
Although our method has achieved excellent results, we discuss certain limitations of our method. Our model renders a patch of pixels in one camera ray, and only use one pixel in the middle of the patch, discarding other pixels, which slightly increases the time cost of rendering the image compared to NeRF. When retaining all pixels in the patch and sampling camera rays using the stride strategy, the rendering speed is significantly improved, but the rendering quality of the image slightly decreased. This indicates that the rendering quality of each pixel in the patch is not the same. Therefore, future directions include improving the rendering quality of all pixels in the patch not just for the center point, and the rendering efficiency of models.

\section{Conclusion}
In this paper, we design a novel torch-NeRF that encourages a single camera ray possessing more contextual information and modeling the relationship among sample points on each camera ray. First, we enlarge the ray perception field to encourage a camera ray aggregating more contextual information through rendering a patch of pixels with a single ray simultaneously. Moreover, we replace the MLP components in neural radiance field models with our proposed distance-aware convolutions along rays, which enhance the feature interaction among sample points on the same camera ray. We carry out extensive experiments on different scenes with complex background and prove that our proposed method achieves significant improvements.

\bibliographystyle{cas-model2-names}

\bibliography{cas-refs}

\end{document}